# Interpreting Basis Path Set in Neural Networks

Juanping Zhu[1] Qi Meng[2] Wei Chen[2] Zhi-ming Ma[3]

1.Yunnan University, 2. Microsoft Research Asia, 3. Chinese Academy of Science

**Abstract:** Based on basis path set, G-SGD algorithm significantly outperforms conventional SGD algorithm in optimizing ReLU neural networks. However, how the inner mechanism of basis paths work remains mysterious. From the aspect of graph theory, this paper defines basis path, investigates structure properties of basis paths in regular fully connected neural network and interprets the graph representation of basis path set. Moreover, we propose hierarchical algorithm HBPS to find basis path set $B$ in fully connected neural network by decomposing the network into several independent and parallel substructures. Algorithm HBPS demands that there doesn't exist shared edges between any two independent substructure paths.

## 1 Introduction

Recently there has been a surge of interest in high interpretability in both theory and practice of neural networks, and more and more researchers tend to develop models with interpretable knowledge representations [1,2,3]. Conventional stochastic gradient descent (SGD) algorithm trains neural networks in the vector space of weights [4], which leads to mismatch during optimizing ReLU neural networks in positively scale-invariant space [5]. [6] proposed the basis path norm in G-space. To handle this mismatch, Meng et al. [7] proposed to find basis path set defined in G-space and optimize the value of basis paths of neural networks by stochastic gradient descent algorithm. So the G-space has lower dimension under the positive scaling group and the experiments turn out that G-SGD significantly outperforms conventional SGD algorithm in optimizing ReLU networks.

In spite of good performance, the basis path set strategy remains mysterious in the training of neural networks of ReLU, like what was mentioned in most recent neural network interpretation in [1,8,9]. Except the final network output and proof of G-space in ReLU network, it is difficult for people to understand how the inner mechanism of basis paths works. How does basis path set help achieve superior performance in the training of ReLU networks. What kind of properties does the basis path set have? How does the basis path set work in regular neural network? To the end of fully understanding how the basis paths work, basis path is defined in regular fully connected neural network, attractive structure properties of basis paths are investigated and the graph representation of basis path set is interpreted in this paper. All research in the paper is undertaken in the aspect of graph theory. Moreover, we propose the hierarchical idea to decompose the structures of neural networks. Hierarchical idea occasionally appears in the algorithm designing in combinatorial optimization [10,11,12,13]. Hierarchical optimization usually decomposes optimization problem into two or more sub-problems, which has its own subjective and constrains [10,11]. Our paper decomposes the fully connected neural network structure into several simple, independent and parallel substructures and finds basis path set for each independent substructure. This hierarchical algorithm doesn't levy any constraints about how the edges skip over layers, but requires no shared edges between any two independent substructure paths in the target neural network.

The contributions of this paper are summarized as follows. 1) Basis path in neural network is defined from the aspect of graph theory. 2) Interpretation of basis path set in neural networks displays elegant graphic structure property. 3) Hierarchical idea finds basis path set for fully connected neural network. The top level of the hierarchical algorithm is to find the maximal independent substructures of a given neural network, which can offer one novel way to explore the structure of fully connected neural network. The lower level of the algorithm is to find basis paths for each independent substructure. The investigation in this paper can help us in training neural networks and can help to explore how neural network works with basis paths.

## 2 Basic Knowledge

### 2.1 Basic definitions and basic operations on the neural network graph

Fully connected neural network is a $L$-layer multi-layer perceptron with weighted edges that can skip over different layers, shown as Fig 1. We denote $i$-th node in $l$-th layer as $O_i^l$ and the node set of the $l$-th layer as $O^l$. We use $(O_i^l, O_{i'}^{l+j})$ to denote the directed edge between layer $l$ and layer $l+j$, where $1 \leq j \leq L-l$, $1 \leq i \leq |O^l|$ and $1 \leq i' \leq |O^{l+j}|$. This paper would interpret neural network from the angle of graph theory. Neural network is described as a triple graph $G = (V, E, w)$, where the finite node set $V = O^0 \cup \ldots \cup O^l \ldots \cup O^L$ comprising nodes from all layers in neural network $G$ and finite edge set $E = \{(O_i^l, O_{i'}^{l+j}) | 0 \leq l \leq L-1$ and $1 \leq j \leq L-l\}$ consisting of all directed fully connected edges between different layers. $w(O_i^l, O_{i'}^{l+j})$ is the weight of edge $(O_i^l, O_{i'}^{l+j}) \in E$, and $m = |E|$ is the number of edges in graph $G$. Neural network is represented as a directed acyclic graph $G$.

**Definition 1** (path) [14] A walk in graph $G$ is an alternating sequence $W = v_1 e_1 v_2 e_2 v_3 \ldots v_{k-1} e_{k-1} v_k$ of nodes $v_i$ and edges $e_j$ from $G$ such that the tail of $e_j$ is $v_j$ and the head of $e_j$ is $v_{j+1}$ for every $j \in \{1, \ldots, k-1\}$. If the nodes of $W$ are distinct, $W$ is a path.

**Definition 2** ($r$-layer-skip edge) If the edge connects the node from the $l$-th layer to the node of the $l + (r+1)$-th layer and skips over $r$ layers, we denote this edge as $r$-layer-skip edge, where $r = 0, 1, \ldots, L-l-1$.

As shown in Fig. 1, 0-layer-skip edge without any layer skipping is solid and the edge skipping over at least one layer such as 1-layer-skip edge is dashed. Some basic operations on the edge are defined in graph $G$, such as edge removal, edge addition and edge swap [14,15,16]. If edge $e \in E(G)$, the removal of the edge is defined as $G - e := (V(G), E(G) \setminus \{e\})$. The addition of a new edge is define as $G + e := (V(G), E(G) \cup \{e\})$[14,15,16]. When we remove one edge $e$ from graph $G$ and add a new edge $e'$ to graph $G$, we call this procedure $G - e + e'$ as an edge swapping. In order to investigate the properties of paths in neural networks, the path operations are specifically defined as follows.

**Definition 3** (path addition to a graph) Given a graph $H$ and a path $p$, we denote the path addition by $H + p$ with $V(H + p) = V(H) \cup V(p)$ and $E(H + p)$ being the disjoint union of $E(H)$ and $E(p)$ (Parallel edges [15,16] may arise).

**Definition 4** (path removal from a graph) Given a graph $H$ and one path $p$ with $E(p) \subseteq E(H)$, the removal of the path $p$ from the graph $H$ is defined as $H - p$ with $E(H - p) = E(H) \setminus E(p)$ and $V(H - p) = V(H) \setminus \{v \in V(H) | v$ is an isolated vertex after $E(H) \setminus E(p)\}$. (Note that graph $H$ can be a multi-graph [15,16] with parallel edges)

**Definition 5** (addition of two paths) The addition of two paths is defined as $p_1 + p_2$ with $V(p_1 + p_2) = V(p_1) \cup V(p_2)$ and $E(p_1 + p_2)$ being the disjoint union of $E(p_1)$ and $E(p_2)$ (Parallel edges may arise and usually path operation $p_1 + p_2$ doesn't form a new path).

### 2.2 Independent path set

Fully connected neural network $G$ has some special graph structure, which can be seen as the space of paths starting from the input layer and ending at the output layer. Graphs are a fundamental combinatorial structure [16] and there must be some unique structure properties to explore in neural network $G$ from the angle of graph theory.

For the simplicity, the symbol $O_{i^*}^l$ is denoted for some node without specified position in the $l$-th layer in the following section. Let $p = (O_{i^*}^0, O_{i^*}^{1'}, ..., O_{i^*}^{j'} ..., O_{i^*}^L)$ be the path starting from the input layer $O^0$ to the output $O^L$ passing through from several hidden nodes $O_{i^*}^{1'}$, $O_{i^*}^{2'},..., O_{i^*}^{j'}$, where $0 < 1' < 2'...< j' < L$. Note that path $p$ can jump or skip over several hidden layers and the number of skipped layers may be not uniform. Let $P = \{(O_{i^*}^0, O_{i^*}^{1'}, ..., O_{i^*}^{j'} ..., O_{i^*}^L) | 0 < 1' < 2'...< j' < L\}$ be the path set consisting of all paths from the input layer to the output layer in network $G$. Thus the cardinality of set $P$ is rather huge [7]. We hope to find some subset $B \subseteq P$, from which we can reach any path $p \in P$ through multiple path operations within path subset $B$ like linearly independent basis vectors. So cardinality of subset $B$ is substantially far smaller than $P$[7]. This paper defines the independent path and basis path set in the aspect of graph theory.

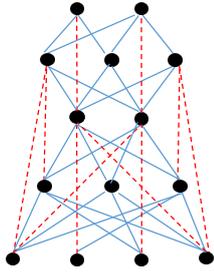 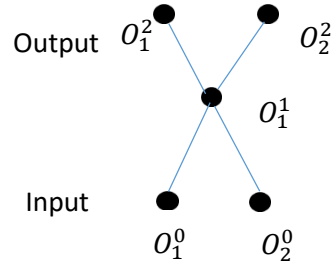

Fig. 1 Neural network with layer skipping     Fig. 2 Example of independent paths

**Definition 6** (independent path set) Given path set $B$, if there exists one path $p \in B$ and another path $q \in B \setminus \{p\}$ such that we can reach path $p$ from path $q$ through finite steps of path addition and path removal within the path set $B$, we call path set $B$ is dependent. Otherwise, we call path set $B$ is independent.

In other words, if path set $B$ is an independent path set, for any path $p \in B$, we couldn't reach path $p$ through operations of addition and removal on any combination of paths in $B \setminus \{p\}$. If path set $B$ is a dependent path set, there exists one path $p$ and path subset $\{q_1, ..., q_s\} \subseteq B \setminus \{p\}$ such that we can get path $p$ through finite times of path operations on path set $\{q_1, ..., q_s\}$. We say the path $p \in B$ can be represented by the path set $B \setminus \{p\}$.

In Fig. 2, path $p_1 = (O_1^0, O_1^1, O_1^2)$, path $p_2 = (O_2^0, O_1^1, O_2^2)$, $p_3 = (O_1^0, O_1^1, O_2^2)$, and $p_4 = (O_2^0, O_1^1, O_1^2)$. Let path set $B = \{p_1, p_2, p_3\}$, where subset $\{p_1, p_2\}$ covers all edges in $B$. Because neither path $p_2$ nor path $p_3$ contains edge $(O_1^1, O_1^2)$, so we couldn't reach path $p_1$ from path operations in set $B \setminus \{p_1\}$. And so on, we couldn't get the path $p_2$ from $\{p_1, p_3\}$ and path $p_3$ from $\{p_1, p_2\}$. Hence, path set $B$ is an independent set. However, if let path set $B = \{p_1, p_2, p_3, p_4\}$, $S$ is an dependent path set for $p_4 = p_1 + p_2 - p_3$. According to Def. 3,4,5,    $E(p_1 + p_2 - p_3) = \{(O_1^0, O_1^1), (O_1^1, O_1^2), (O_2^0, O_1^1), (O_1^1, O_2^2)\} \setminus \{(O_1^0, O_1^1), (O_1^1, O_2^2)\}$ $= \{(O_2^0, O_1^1), (O_1^1, O_1^2)\} = E(p_4)$.

### 2.3 Basis path set--- maximal independent path set

**Definition 7** (maximal independent path set) A path set $B$ of neural network graph $G$ is maximally independent, if including any other path $p^* \in P$ would make $B \cup p^*$ dependent.

**Definition 8** (basis path set) Given the path set $P$ of neural network $G$, the path subset $B \subseteq P$ is said to be the basis path set if path set $B$ is independent and every path $p \in P$ starting from the input layer to output layer can be represented by path set $B$ by the finite path operations. Path $p \in B$ is called basis path.

Hence, for the maximal independent path set $B$, any path $p \in P$ of neural network $G$ can be reached as a combination of elements of $B$ by the finite number of path operations. A basis path set $B$ of neural network $G$ is a maximal independent subset of $P$. In other words, a superset of a basis path set is dependent. A maximal independent subset of path set $P$ of neural network $G$ is a basis path set.

## 3 Basis Path Representation of Neural Network

In this section we will study from the perspective of graph theory about how we represent any path $p \in P$ in neural network $G$ within basis path set $B$ through finite path operations and about how we find the basis path set $B$ in the fully connected network graph $G$ without any edge-skipping. Later the weights on the paths in ReLU network [7] can be easily manipulated through the path operations within the basis path set. In order to represent neural network through basis path set $B$, the investigations on the properties of basis path set $B$ in graph $G$ should be undertaken first.

**Lemma 1** The basis path set $B$ must cover all edges of the neural network graph $G$.
**Proof:** Assume edge $e$ is not covered by the basis path set $B$. We can find one path $p = (O_{i^*}^0, \ldots, O_{i^*}^L)$ from the input layer to the output layer passing through edge $e$. Then there is no way for us to reach path $p$ by the operations of adding and removing paths within $B$, because no path in $B$ contains edge $e$.

Based on Lemma 1, Lemma 2 considers the basis path set $B$ of the unbalanced network $G$ without edge skipping. Here unbalanced network $G$ is defined as a network such that the number of nodes in each layer is not necessary to be the same. This lemma will tell us how to construct the independent path set $B$ in a recursive way. After independent path set is proved, we will discuss that we can reach any path $p \in P/B$ by starting from $p_0 \in B$ and taking path operations within independent path set $B$. Hence path set $B$ is a maximal independent path set. During the process of recursive construction of basis paths and of proof of maximal independent path set, the attractive structure properties are explored. These properties are the combinatorial structures companied by the characteristics of independent paths and structures of paths in neural networks.

**Lemma 2** For the fully connected and unbalanced graph $G$ of neural network without any edge skipping over layers, the cardinality of the basis path set $B$ is $m - H$, where $m$ is the number of the edges in graph $G$ and $H$ is the number of the hidden nodes in graph $G$.
**Proof:** For the unbalanced graph $G$, we can follow the idea that first constructs the direct paths and then constructs the cross paths recursively in sub-graph $G(k)$. Here sub-graph $G(k) = (O^k \cup O^{k+1}, E^k)$, where $E^k = \{e \in G | e$ leaves from $k$-th layer and enters $k+1$-th layer $\}$ and $0 \le k \le L - 1$.

First, let $k = 0$ and construct direct paths and cross paths in the sub-graph $G(0)$ with the beginning two layers $O^0$ and $O^1$. If $|O^0| \ge |O^1|$, find $|O^1|$ direct vertex disjoint paths by depth-first searching and let the direct path set be $P_{dir}^{(0)}$. Pick up one node $O_{i'}^1 \in O^1$ randomly and set $P_{dir}^{(0)} = P_{dir}^{(0)} \cup \{(v, O_{i'}^1) | v \in O^0 \setminus V(P_{dir}^{(0)})\}$. If $|O^0| < |O^1|$, find $|O^0|$ direct vertex disjoint paths by depth-first searching and let the direct path set be $P_{dir}^{(0)}$. Let cross path set $P_{cross}^{(0)} = \{e | e \in E^0 \setminus E(P_{dir}^{(0)})\}$, where $E^0 = \{e \in G | e$ leaves from 0-th layer and enters 1-th layer$\}$. So we can see $|P_{dir}(O_i^0)| = 1$ for $i = 1, 2, \ldots, |O^0|$, i.e., only one direct path can pass through each node $O_i^0$ at the 0-th layer. For each node $O_i^0 \in O^0$, classify the direct path and cross paths starting from it, and let the path set $P_{dir}(O_i^0) = \{p \in P_{dir}^{(0)} | $the tail of $p$ is node $O_i^0\}$ and let the path set $P_{cross}(O_i^0) = \{p \in P_{cross}^{(0)} | $the tail of $p$ is node $O_i^0\}$. For each node

$O_i^1 \in O^1$, let path set $P(O_i^1)$ be the path set consisting of both direct paths and cross paths passing through it in $G(0)$. Secondly, construct the direct path set $P_{dir}^{(1)}$ and cross path set $P_{cross}^{(1)}$ in sub-graph $G(1)$ with the 1-th and 2-th layers as described for $G(0)$. In $G(1)$, get path set $P_{dir}(O_i^1)$ and $P_{cross}(O_i^1)$ for each node $O_i^1 \in O^1$. Concatenate the direct path from $P_{dir}(O_i^1)$ with the paths passing each node $O_i^1$ from the lower layer. Path set $P(O_i^1)$ includes all paths reaching node $O_i^1$ from the lower layer. For the only direct path $p_1 \in P_{dir}(O_i^1)$ and all $p_0 \in P(O_i^1)$, do path concatenation as $p_0 + p_1$, which forms a new path. Hence for each node $O_i^1$, the only direct path $p_1 \in P_{dir}(O_i^1)$ is extended to $|P(O_i^1)|$ direct paths. Update direct path set $P_{dir}^{(1)}$. In sub-graph $G(1)$, $P_{cross}(O_i^1)$ includes all cross paths starting from $O_{i_1}^1$. Pick up one path $p^* \in P(O_i^1)$ randomly and extend the cross path set by $p^* + p_1$ for all $p_1 \in P_{cross}(O_{i_1}^1)$. Update cross path set $P_{cross}^{(1)}$. Then gather all paths for set $P(O_i^2)$ passing through each node $O_i^2 \in O^2$.

Thirdly, we can recursively construct the direct path set $P_{dir}^{(k)}$ and cross path set $P_{cross}^{(k)}$ in sub-graph $G(k)$ with the $k$-th and $k+1$-th layers. As described in the discussion of sub-graph $G(2)$, get direct path set $P_{dir}(O_i^k)$ and cross path set $P_{cross}(O_i^k)$ for each node $O_i^k \in O^k$. Let $P_{dir}^{(k)} = \{P_{dir}(O_i^k)|O_i^k \in O^k\}$ and $P_{cross}^{(k)} = \{P_{cross}(O_i^k)|O_i^k \in O^k\}$. Update direct path set $P_{dir}^{(k)}$ by concatenating each direct path $P_{dir}(O_i^k) \in P_{dir}^{(k)}$ with all paths ending at $P(O_i^k)$. Update cross path set $P_{cross}^{(k)}$ by concatenating each cross path $P_{cross}(O_i^k) \in P_{cross}^{(k)}$ with one randomly picked path ending at $P(O_i^k)$ for each node $O_i^k$ at the $k$-th layer, where path set $P(O_i^k)$ includes all direct paths and cross paths reaching node $O_i^k$ from the input layer. Then we can get all paths reaching each node $O_i^{k+1} \in O^{k+1}$ for path set $P(O_i^{k+1})$. Repeat this procedure till $k = L - 1$, and get the updated direct path set $P_{dir}^{(L-1)}$ and cross path set $P_{cross}^{(L-1)}$. Let $B = P_{dir}^{(L-1)} \cup P_{cross}^{(L-1)}$. More details refer to **Subroutine($G$)**.

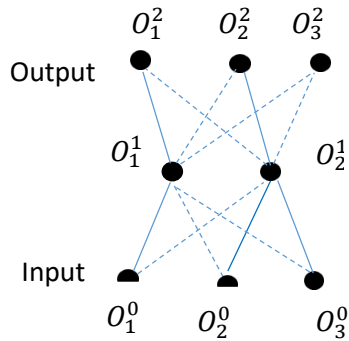

Fig. 3 Constructing independent path

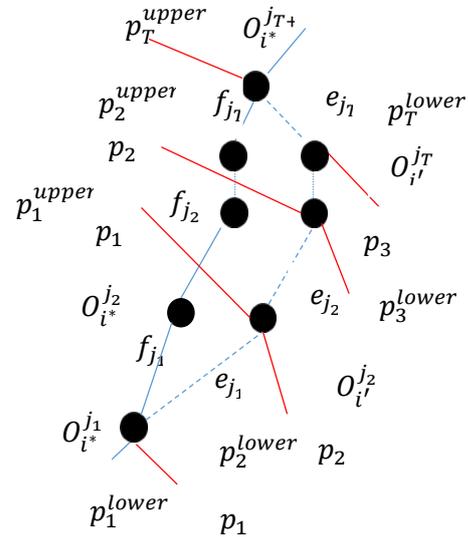

Fig. 4 Proof of basis path set

Fig. 3 is 3-layers unbalanced network. In the first two layers, we can get 2 direct paths $(O_1^0, O_1^1)$ and $(O_2^0, O_2^1)$ and get the third direct path by connecting $O_3^0$ and $O_2^1$. Next, we can get three dashed cross paths. So one solid direct path and two dashed cross paths in path set $P(O_1^1) = \{(O_1^0, O_1^1), (O_2^0, O_1^1), (O_3^0, O_1^1)\}$ reaching node $O_1^1$. Three paths in path set $P(O_2^1) = \{(O_1^0, O_2^1), (O_2^0, O_2^1), (O_3^0, O_2^1)\}$ reaching $O_2^1$. Direct path set $P_{dir}^{(1)} = \{(O_1^1, O_1^2), (O_2^1, O_2^2)\}$ and cross path set $P_{cross}^{(1)} = \{(O_1^1, O_2^2), (O_1^1, O_3^2), (O_2^1, O_1^2), (O_2^1, O_3^2)\}$. The direct path needs accept any path from the lower level. So the concatenated direct path set $P_{dir}(O_1^1) = \{(O_1^0, O_1^1, O_1^2)$, $(O_2^0, O_1^1, O_1^2), (O_3^0, O_1^1, O_1^2)\}$ and $P_{dir}(O_2^1) = \{(O_1^0, O_2^1, O_2^2), (O_2^0, O_2^1, O_2^2), (O_3^0, O_2^1, O_2^2)\}$. Randomly pick path $(O_2^0, O_1^1)$ for $O_1^1$ and get the concatenated cross path set $P_{cross}(O_1^1) = \{(O_2^0, O_1^1, O_2^2), (O_2^0, O_1^1, O_3^2)\}$. We can also get $P_{cross}(O_2^1) = \{(O_2^0, O_2^1, O_1^2), (O_2^0, O_2^1, O_3^2)\}$. Therefore, $P(O_1^2) = \{(O_1^0, O_1^1, O_1^2), (O_2^0, O_1^1, O_1^2), (O_3^0, O_1^1, O_1^2), (O_2^0, O_2^1, O_1^2)\}$, $P(O_2^2) = \{(O_1^0, O_2^1, O_2^2), (O_2^0, O_2^1, O_2^2), (O_3^0, O_2^1, O_2^2), (O_2^0, O_1^1, O_2^2)\}$ and $P(O_3^2) = \{(O_2^0, O_1^1, O_3^2), (O_2^0, O_2^1, O_3^2)\}$.

Now we will use induction on the layer of $k$ to prove that $|B(k)| = m(k) - H(k)$, where $m(k)$ is the number of the edges and $H(k)$ is the number of the hidden nodes in sub-graph $G(k)'$. Here $G(k)'$ is the sub-graph of $G$ from 0-th layer till $k+1$-th layer in graph $G$ and path set $B(k) = P_{dir}^{(k)} \cup P_{cross}^{(k)}$ in $G(k)'$.

Basis step: $k = 0$. Sub-graph $G(0)' = G(0)$ consists of only 0-th and 1-th layers without hidden nodes. $|B(0)| = m(0) - H(0)$ is trivial according to the construction of direct and cross paths. Note that the direct path set and cross path set between only two layers exactly cover all edges.

Induction step: $k \geq 1$. We suppose that the claim $|B(k-1)| = m(k-1) - H(k-1)$ holds for $G(k-1)'$ with layers less than $k+1$. The procedure of updating set $P_{dir}^{(k)}$ and set $P_{cross}^{(k)}$ in $G(k)$ produces totally $\sum_{O_i^k \in O^k} |P_{dir}(O_i^k)| * |P(O_i^k)| + |P_{cross}(O_i^k)|$ paths. According to the assumption, $\sum_{O_i^k \in O^k} |P(O_i^k)| = m(k-1) - H(k-1)$ and $|O^k|$ is the number of nodes in $k$-th layer. Therefore, after updating $\sum_{O_i^k \in O^k} |P_{dir}(O_i^k)| + |P_{cross}(O_i^k)| = m(k) - m(k-1)$, where $m(k) - m(k-1)$ is the number of increased edges after the $k$-th layer is included. As mentioned in the basis step and Lemma 1, the direct path set and cross path set between only any two layers exactly cover all edges between them. According to the rule of path construction in $G(k)$, $|P_{dir}(O_i^k)| = 1$ for any node $O_i^k \in O^k$. Hence,

$|B(k)| = \sum_{O_i^k \in O^k} |P_{dir}(O_i^k)| * |P(O_i^k)| + |P_{cross}(O_i^k)|$
$= \sum_{O_i^k \in O^k} |P_{dir}(O_i^k)| * |P(O_i^k)| + \sum_{O_i^k \in O^k} |P_{cross}(O_i^k)|$
$= \sum_{O_i^k \in O^k} |P_{dir}(O_i^k)| * |P(O_i^k)| + m(k) - m(k-1) - \sum_{O_i^k \in O^k} |P_{dir}(O_i^k)|$
$= \sum_{O_i^k \in O^k} 1 * (|P(O_i^k)| - 1) + m(k) - m(k-1)$
$= m(k-1) - H(k-1) - |O^k| + m(k) - m(k-1) = m(k) - H(k)$.

Repeat this induction step till $k = L - 1$. By the induction hypothesis, $|B(L-1)| = m(L-1) - H(L-1)$. Sub-graph $G(L-1)'$ is graph $G$, $m(L-1)$ is the number $m$ of the edges in $G$ and $H(L-1)$ is the number $H$ of the hidden nodes in $G$. Hence, $|B| = m - H$.

Next, we will discuss $B(L-1)$ is a basis path set for graph $G$. First, all edges in $G$ are covered by $B(L-1)$, because every edge between any two consecutive layers is either direct path set or cross path. Secondly, path set $B(L-1)$ is independent, because no path $p \in B(L-1)$ can be represented by $B(L-1)/p$ for the recursive way of constructing direct path and cross path. Then, we will prove that any path $p \in P/B(L-1)$ in graph $G$ can be represented by the elements of $B(L-1)$ through finite path operations based on edge swapping. For simplicity, suppose that any edge swapping only takes place within the same layers and all edge swapping is done consecutively in succession. Some other cases can be analyzed in the same way. Given any target path $p = (O_{i^*}^0, e_0, ..., O_{i^*}^{j_1}, e_{j_1}, ..., O_{i^*}^{j_t}, e_{j_t}, ..., O_{i^*}^{j_T}, e_{j_T}, O_{i^*}^{j_{T+1}} ..., O_{i_L}^L) \in P/B(L-1)$, we can find one path $p_0 \in B(L-1)$ such that $p$ and $p_0$ start from the same input $O_{i^*}^0$ and both paths share as many edges as possible after node $O_{i^*}^0$. Let $O_{i^*}^{j_1}$ be the layer node where paths $p$ and $p_0$ separate and let $O_{i^*}^{j_{T+1}}$ be the node where paths $p$ and $p_0$ merge again. Assume $p_0 = (O_{i^*}^0, e_0, ..., O_{i^*}^{j_1}, f_{j_1}, O_{i'}^{j_2} ..., O_{i'}^{j_t}, f_{j_t}, ..., O_{i'}^{j_T}, f_{j_T}, O_{i^*}^{j_{T+1}} ..., O_{i^*}^L)$, where node $O_{i'}^{j_t} \neq O_{i^*}^{j_t}$ and $2 \leq t \leq T \leq L - 1$. Find paths $p_1,\ldots, p_t,\ldots,p_T$ in path set $B(L-1)$ such that path $p_t$ takes only edge $e_{j_t}$ at $O_{i*}^{j_t}$, $1\leq t \leq T \leq L-1$. For example, $p_t = p_t^{lower} + e_{j_t} + p_t^{upper}$ as shown in Fig. 4. The previous discussion turns out that only one direct path can leave each hidden node $O_{i*}^{j_t}$ in the $j_t$-th layer. In Fig. 4, the blue solid path is the starting path $p_0$ and blue dashed path is the target $p$. Red path $p_t$ and $p_{t-1}$ pass $O_{i*}^{j_t}$, and $e_{j_{t-1}} \notin p_t$. So $e_{j_t}$ is a cross path within $j_t$-th and $j_{t+1}$-th layers, otherwise path $p_{t-1}$ can stretch longer. Set $p_1' = p_1^{lower} + (f_{j_1},\ldots, f_{j_T}) + p_T^{upper}$. Sub-path $p_1^{lower}$ and $f_{j_1}$ share node $O_{i*}^{j_1}$ for $e_{j_1}$ and cross path $e_{j_1}$ takes the randomly selected path $p_1$. If $f_{j_1}$ is a cross path, it must follow $p_1^{lower}$ according to the rule of construction. So $f_{j_1}$ is a direct path and admissible for all paths passing through $O_{i*}^{j_1}$. In the meantime, paths $p_0$ and $p_T$ share the sub-path from $O_{i*}^{j_{T+1}}$ till $O_{i_L}^L$, which turns out that $e_{j_{T+1}}$ is a direct path and admissible for path set $P\left(O_{i*}^{j_{T+1}}\right)$. And $(f_{j_1},\ldots, f_{j_T})$ is part of path $p_0$. Therefore, $p_1'$ forms a path in $B(L-1)$, because edge $f_{j_1}$ is a direct path and $e_{j_{T+1}}$ is a direct path too. Let $p_t - e_{j_t} = p_t^{lower} + p_t^{upper}$. $p_t^{lower}$ and $p_{t-1}^{upper}$ share node $O_{i*}^{j_t}$. Because path $e_{j_t}$ is a cross path and $p_t$ passes it, so path $p_t^{lower}$ is the randomly selected path. Moreover $p_{t-1}$ passes $O_{i*}^{j_t}$ too, so $f_{j_t}$ must be a direct path and admissible for $P\left(O_{i*}^{j_t}\right)$. Therefore, $p_t' = p_t^{lower} + p_{t-1}^{upper}$ forms a path in $B(L-1)$ for $2 \leq t \leq T$. Finally,

$p = p_0 + e_{j_1} - f_{j_1} + \cdots + e_{j_t} - f_{j_t} \ldots + e_{j_T} - f_{j_T} = p_0 + p_1 \setminus \left(p_1^{lower} + p_1^{upper}\right) - f_{j_1} + \cdots + p_t \setminus \left(p_t^{lower} + p_t^{upper}\right) - f_{j_t} \ldots + p_T \setminus \left(p_T^{lower} + p_T^{upper}\right) - f_{j_T} = p_0 + p_1 \ldots + p_t \ldots + p_T - \left(p_1^{lower} + f_{j_1} \ldots + f_{j_T} + p_T^{upper}\right) - \left(p_1^{upper} + p_2^{lower}\right) \ldots - \left(p_{t-1}^{upper} + p_t^{lower}\right) \ldots + \left(p_{T-1}^{upper} + p_T^{lower}\right)$ $= p_0 + p_1 \ldots + p_t \ldots + p_T - p_1' \ldots - p_t' \ldots - p_T'$.

Therefore, any given target path $p \in P/B(L-1)$ can be represented by finite steps of path operations in $B(L-1)$. So $B = B(L-1)$ is a basis path set with cardinality of $m - H$. ∎

## 4 Hierarchical Algorithm to find the basis path set

The proof of Lemma 2 investigates uniquely excellent properties of direct paths and cross paths when constructing the basis paths recursively in neural network $G$. These properties allow any path $p \in P$ can be represented by basis path set $B$, in graph theory. However, this neural network $G$ is a graph without skipping-edges. In this section, we will explore the structure of fully connected graph $G$ with any possible edge-skipping first and then offer one hierarchical idea to find basis path set $B$ in such neural network without shared layers between any independent substructures.

**Definition 9** (structure path) Given fully connected neural network $G$, if all paths in $P$ passes through the same layers consecutively, any path $p \in P$ can be called as the structure path of neural network $G$, since such path can express the structure information of $G$.

Here the structure information of neural network $G$ means how the path $p \in P$ passes through layers from the input layer to the output layer, *i.e.*, how the path $p \in P$ skips over layers. Especially any path $p \in P$ in fully connected neural network $G$ without edge skipping is a structure path, because all paths in $P$ pass the layers homogenously.

**Property 1** If fully connected neural network graph $G$ has a structure path $p$, then graph $G$ is homogenous, *i.e.,* all paths from the input layer to the output layer homogenously pass through the same layers consecutively as $p$.

As we know, the role of every node in the $l$-th layer in the fully connected network $G$ is equivalent, so any node $O_{i*}^l$ can be the representative of the rest nodes in the $l$-th layer. This motivates us that we can decompose the complicated structure of graph $G$ into several distinct and simple substructures and each substructure has its unique structure path. For each substructure, we induce the corresponding sub-graph by including all layers passed by the structure path $p$ and including all edges connecting to these layers in the order of the layers passed by $p$ in graph $G$.

**Definition 10** (substructure path) Let $V^S = \{O_{i*}^l | l = 0, \ldots, L\}$, where $O_{i*}^l$ is the random node selected from the $l$-th layer of network $G$. Let $E^S = \{(O_{i*}^k, O_{i*}^l) \in E | 0 \leq k < l \leq L\}$. In this simplified sub-graph $G^S = (V^S, E^S)$ with only one node at each layer, we can get all paths starting from $O_{i*}^0$ to $O_{i*}^L$ by breadth-first search. Denote this path set as $P^S$ and each path $p \in P^S$ is called substructure path.

**Lemma 3** Path set $P^S$ covers all substructure information in fully connected neural network $G$.

**Proof:** Neural network $G$ is fully connected, any node in the $l$-th layer can represent the rest nodes in this layer, when discussing the structure in formation. Since we find all paths from node $O_{i*}^0$ to node $O_{i*}^L$ by breadth-first search, so it is trivia that path set $P^S$ covers all substructure information in $G$. ∎

In fully connected network in Fig. 5, sub-graph $G_{blue}$ consists of all blue solid edges and sub-graph $G_{red}$ consists of all red dashed edges. Path $(O_1^0, O_1^1, O_1^2)$ and path $(O_1^0, O_1^2)$ are two substructure paths and corresponding sub-graphs are $G_{blue}$ and $G_{red}$.

By taking the idea of adjacent matrix in graph theory, we can express each substructure path $p_i \in P^S$ as a 0-1 vector $\alpha_i$ with $(L+1)^2$ elements. First, construct one $(L+1) \times (L+1)$ adjacency matrix $M_i$ such that its element $M_i(j, l)$ is one when there is edge from $j$-th layer to the $l$-th layer and zero when there is no edge in path $p_i$, where $j = 0, 1, \ldots, L$ and $l = 0, 1, \ldots, L$. Obviously, matrix $M_i$ is an upper triangle and sparse matrix. Then we reshape matrix $M_i$ into one $1 \times (L+1)^2$ vector $\alpha_i$, by letting $\alpha_i = [M_i(0, :), \ldots, M_i(j, :), \ldots, M_i(L, :)]$, where $M_i(j, :)$ is the $j+1$-th row of matrix $M_i$ for the edges leaving $j$-th layer. So the structure of $p_i$ determines the vector $\alpha_i$, and we call this $(L+1)^2$-dimensional vector $\alpha_i$ as substructure path vector of $p_i$. In Fig. 5, the vector of substructure path $(O_1^0, O_1^1, O_1^2)$ is $(0,1,0,0,0,1,0,0,0)$ and the vector of $(O_1^0, O_1^2)$ is $(0,0,1,0,0,0,0,0,0)$. Given the substructure path set $P^S$ in $G^S$ and corresponding substructure path vector set $\{\alpha_i | i = 1, 2, \ldots, |P^S|\}$, one substructure path vector $\alpha_i$ maps to one substructure path $p_i$, and one substructure path $p_i$ decides one vector $\alpha_i$. According to the previous definition of path operation, there is one one-one mapping from the linear combination of substructure path vectors in $\{\alpha_i | i = 1, 2, \ldots, |P^S|\}$ to the path operation in $P^S$, i.e., $p_t = p_{r_1} \ldots + p_{r_j} \ldots + p_{r_d} - p_{s_1} \ldots - p_{s_h} \ldots - p_{s_m}$ corresponds to $\alpha_t = \alpha_{r_1} \ldots + \alpha_{r_j} \ldots + \alpha_{r_d} - \alpha_{s_1} \ldots - \alpha_{s_h} \ldots - \alpha_{s_m}$, where $r_j \in \{1, 2, \ldots, |P^S|\}$ and $s_h \in \{1, 2, \ldots, |P^S|\}$.

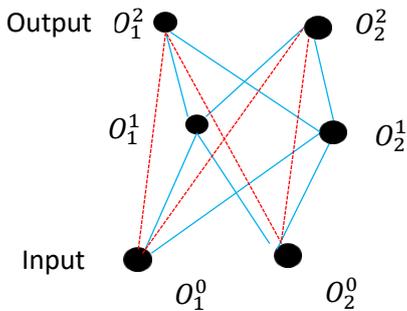

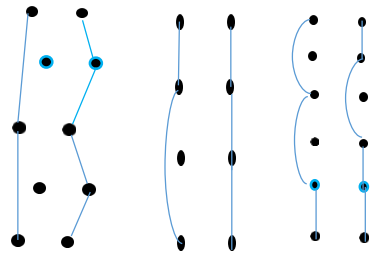

Fig. 6 Substructure paths

Fig. 5  Substructures of neural network  Fig. 6(a)   Fig. 6(b)   Fig. 6(c)

**Definition 11** (maximal independent substructure path set) Given fully connected neural network $G$ and substructure path set $P^S$. If subset $\{\alpha_i\}_{ind}$ of 0-1 substructure path vector set $\{\alpha_i\}$ with respect to $P^S$ is linearly independent, the corresponding substructure path set $P^S_{ind}$ is called to be independent, and the structures of induced sub-graphs from the corresponding paths in $P^S_{ind}$ are called independent substructures. Otherwise, $P^S_{ind}$ is called to be dependent. When subset $\{\alpha_i\}_{ind}$ is maximal linearly independent, we call the corresponding $P^S_{ind}$ as maximal independent substructure path set in graph $G$, where $|P^S_{ind}|$ is the number of substructure paths in $P^S_{ind}$.

Therefore, we can get maximal independent substructure path set $P^S_{ind}$ by numerical linear algebra method. One substructure path within $P^S_{ind}$ couldn't be linearly represented by the rest of substructure paths in $P^S_{ind}$. The substructure path $p \in P^S \setminus P^S_{ind}$ can be represented by $P^S_{ind}$ by path operations and vector linear combination. In Fig. 5, these two 9-dimensional vectors form maximal linearly independent substructure path set.

**Property 2** Maximal independent substructure path set $P^S_{ind}$ can represent the whole structure information of neural network $G$.

Note that the sub-graphs of corresponding substructure paths may overlap at some layers but not all layers and each sub-graph with homogenous layer-passing structure can be treated like a fully connected graph without any skipping edges. Therefore, hierarchical Algorithm HBPS is proposed to find the basis path set in regular fully connected graph with edge-skipping. First, the upper level of the hierarchical idea is to decompose the complicated structure of neural network $G$ into $|P^S_{ind}|$ maximal independent substructures, which are simple and whose induced sub-graph can be treated as network without edge skipping. Any substructure path $p \in P^S \setminus P^S_{ind}$ can be represented by $P^S_{ind}$. Substantially the upper level of algorithm is seeking the inter-structure independence. Then, the lower level of the hierarchical idea will find the basis path set $B_r$ for each sub-graph $G_r$ induced by $p_r \in P^S_{ind}$ in parallel. For each $p_r$, we construct sub-graph $G_r$ of $G$ by taking all nodes from the layers through which $p_r$ passes and taking all edges connecting these nodes in $G$ in the same order as $p_r$. In sub-graph $G_r$, call **Subroutine**($G_r$) which is recursively designed to find the basis path set for regular unbalanced graph described in Lemma 2. We must emphasize that Algorithm HBPS can be only applied in the network without shared layers between any two substructures such as Fig. 6(a). Two substructure paths in Fig. 6(b) share the edge from the 2-th layer to 3-th layer, and two substructure paths in Fig. 6(c) share the edge from the 0-th layer to 1-th layer.

**Algorithm HBPS**
**Input:** Fully connected unbalanced neural network $G = (V, E)$ with $L$ layers
**Output:** Path set $B$ of neural network $G$

*% Step 1. (The upper level) %*
**For** $l = 0, \dots, L$ do
    Pick up one node $O^l_{i*}$ randomly at the $l$-th layer of graph $G$.
**End For**
Set the node subset $V^S = \{O^l_{i*} | 0 \leq l \leq L\}$ and edge subset $E^S = \{(O^j_{i*}, O^l_{i*}) \in E | O^j_{i*} \in V^S, O^l_{i*} \in V^S, 0 \leq j < l \leq L\}$. Let $P^S$ be the path set for all paths starting from node $O^0_{i*}$ to node $O^L_{i*}$ by breadth-first search in sub-graph $G^S = (V^S, E^S)$.
**For** $i = 1: |P^S|$ do
    For path $p_i \in P^S$, construct adjacency matrix $M_i$,
    $M_i(j, l) = \begin{cases} 1, & \text{if there is edge from } j-\text{th layer to the } l-\text{th layer} \\ 0, & \text{otherwise} \end{cases} \quad \begin{matrix} j = 0,1,\dots,L \\ l = 0,1,\dots,L \end{matrix}$
    Let substructure path vector $\alpha_i = [M_i(0,:), \dots, M_i(j,:), \dots, M_i(L,:)]$.
**End For**

Find the maximal linearly independent subset $\{\alpha_i\}_{ind}$ of $\{\alpha_i\}$ and its corresponding substructure path set $P^S_{ind}$ by numerical linear algebra method.

**For** $r= 1:|P^S_{ind}|$ do
    **For** $i = 1:|P^S_{ind}|$ do
        If $E(p_i) \cap E(p_r) \neq \emptyset$, output 'There exist shared edges between two independent substructure paths' and exit.
    **End**
**End**

*%Step 2. (The lower level) %*
**For** $r= 1:|P^S_{ind}|$ do
    For substructure path $p_r \in P^S$, let node subset $V_r = \{O^l \subset V | O^l_{i*} \in p_r, 0 \leq l \leq L\}$ and $E_r = \{(O^l_i, O^k_{i'}) \in E | O^l \in V_r, O^k \in V_r, i = 1,2,...,|O^l|, i' = 1,2,...,|O^k|\}$. Let sub-graph $G_r = (V_r, E_r)$. *% Construct sub-graph $G_r$ with the same structure as substructure path $p_r$%*
    Run **Subroutine** $(G_r)$ on $G_r$ and output basis path set $B_r$ of $G_r$.
**End For**
Output $B = \bigcup_{r=1}^{|P^S_{ind}|} B_r$.

**Subroutine**$(G)$:
*% This subroutine is to calculate the basis path set on graph without any edge skipping over layers*
**Input:** Fully connected neural network graph $G = (V, E)$ without any edge skipping over layers.
**Output:** Basis path set $B$ in graph $G$
**For** $k = 0: L - 1$ do
    Let $E^k = \{e \in G | e$ leaves from $k$-th layer and enters $k+1$-th layer $\}$.
    *% Step 1. Construct the direct path set.*
    Let sub-graph $G(k) = (O^k \cup O^{k+1}, E^k)$.
    **If** $|O^k| \geq |O^{k+1}|$ do
        Find $|O^{k+1}|$ vertex disjoint paths by depth-first searching and let the path set be $P^{(k)}_{dir}$. For $v \in O^k \setminus V(P^{(k)}_{dir})$, pick up one node $O^{k+1}_{i'} \in O^{k+1}$ randomly and construct path $(v, O^{k+1}_{i'})$. Set $P^{(k)}_{dir} = P^{(k)}_{dir} \cup (v, O^{k+1}_{i'})$. *% $P^{(k)}_{dir}$ is direct path set.*
    **Else** do
        Find $|O^k|$ vertex disjoint paths by depth-first searching and let the path set be $P^{(k)}_{dir}$.
    **End If**
    *% Step 2. Construct the cross path set. %*
    Set cross path set $P^{(k)}_{cross} = E^k \setminus E(P^{(k)}_{dir})$.
    **For** $i = 1:|O^k|$
        Let the path set $P_{dir}(O^k_i) = \{p \in P^{(k)}_{dir} |$ the tail of $p$ is node $O^k_i\}$ and $P_{cross}(O^k_i) = \{p \in P^{(k)}_{cross} |$ the tail of $p$ is node $O^k_i\}$. *% Classify direct paths and cross paths for each node $O^k_i \in O^k$.*
    **End For**
    *% Step 3. Concatenate the direct paths and cross paths from the $k-1$-th layer. %*
    **If** $k \neq 0$ do *% If $k = 0$, there is no concatenation for any path and go to Step 4 directly.*
        **For** $i = 1:|O^k|$ do
            Let $P_{dir}(O^k_i) = \{p_0 + p_1 | p_1 \in P_{dir}(O^k_i), p_0 \in P(O^k_i)\}$ for node $O^k_i \in O^k$.
            *% form $|P(O^k_i)|$ direct paths%*
            Select one path $p^* \in P(O^k_i)$ randomly and let $P_{cross}(O^k_i) = \{p^* + p_1 | p_1 \in P_{cross}(O^k_i)\}$. *% extend all cross paths %*
        **End For**
        Update $P^{(k)}_{dir} = \bigcup_{O^k_i \in O^k} P_{dir}(O^k_i)$ and $P^{(k)}_{cross} = \bigcup_{O^k_i \in O^k} P_{cross}(O^k_i)$.
    **End If**
    *% Step 4. Classify the paths for the nodes in the $k+1$-th layer. %*

   **For** $i = 1: |O^{k+1}|$ do
     Let the path set $P(O_i^{k+1}) = \{p \in P_{dir}^{(k)} \cup P_{cross}^{(k)} \mid$ the head of $p$ is node $O_i^{k+1}\}$.
   **End For**
  **End for**
Output basis path set $B = P_{dir}^{(L-1)} \cup P_{cross}^{(L-1)}$.    ∎

**Theorem 1** Given a fully connected neural network graph $G$, **Algorithm HBPS** can find a basis path set $B = \cup_{r=1}^{|P_{ind}^S|} B_r$. The cardinality of set $B$ is $\sum_{r=1}^{|P_{ind}^S|}(m_r - H_r)$, where $P_{ind}^S$ is the maximal independent substructure path set of graph $G$ and $B_r$ is the basis path set of induced sub-graph $G_r$ with respect to $p_r \in P_{ind}^S$. $m_r$ is the number of total edges and $H_r$ is the number of total hidden nodes in sub-graph $G_r$.

**Proof:** The upper level of Algorithm HBPS finds the maximal independent substructure path set $P_{ind}^S$ of neural network $G$. For any independent substructure path $p_r \in P_{ind}^S$, it couldn't be represented by $P_{ind}^S \setminus \{p_r\}$. Any substructure path $p \in P^S / P_{ind}^S$ can be represented by $P_{ind}^S$. The lower level of Algorithm HBPS finds the basis path set $B_r$ in sub-graph $G_r$ in parallel. In order to prove the path set $B = \cup_{r=1}^{|P_{ind}^S|} B_r$ is a basis path set, we need to prove $B$ is an independent path set and every path $p \in P \setminus B$ can be represented by the elements of $B$.

First, for each sub-graph $G_r$ represented by path $p_r \in P_{ind}^S$, we can get basis path set $B_r$ by **Subroutine**($G_r$). All paths in set $B_r$ are independent. Secondly, for independent substructure paths $p_r \in P_{ind}^S$ and $p_s \in P_{ind}^S$ ($s \neq r$), any path $p \in B_s$ couldn't be represented by $B_r$ for different substructure. And any path $p \in B_s$ couldn't be presented by $\cup_{r=1, r\neq s}^{|P_{ind}^S|} B_r$. Assume path $p_s^* \in B_s$ can be represented by some path $p_{r_1}', p_{r_2}', \ldots, p_{r_i}', \ldots, p_{r_d}'$, where we require all $r_i$ are distinct for $p_{r_i}' \in B_{r_i}$, $r_i \in \{1, 2, \ldots, |P_{ind}^S|\}$ and $i = 1, \ldots, d$. Since Property 1 indicates that path $p_{r_i}'$ has the same structure of $p_{r_i}$, we reduce the path $p_{r_i}'$ to the substructure path $p_{r_i}$ and reduce $p_s^*$ to $p_s$. So $p_s$ can be represented by $\{p_{r_i} \mid r_i \in \{1, 2, \ldots, |P_{ind}^S|, i = 1, \ldots, d\}$, which contradicts the claim that $P_{ind}^S$ is maximal independent.

Furthermore, we will prove that any path $p \in B_s$ couldn't be represented by $\cup_{r=1}^{|P_{ind}^S|} B_r$. Note the constraint of Algorithm HBPS is that $E(p_i) \cap E(p_r) = \emptyset$ for any $p_i \in P_{ind}^S$ and $p_r \in P_{ind}^S$ while $i \neq r$. Because each $p_r \in P_{ind}^S$ has unique structure, if $p \in B_s$ can be represented by $\cup_{r=1}^{|P_{ind}^S|} B_r$, it must be $p - p_{s,1} + p_{r_1,1} - p_{r_1,2} + \cdots p_{r_i,1} - p_{r_i,2} + \cdots + p_{r_d,1} - p_{r_d,2} = 0$, where $p_{s,1} \in B_s, p_{r_i,1} \in B_{r_i}$ and $p_{r_i,2} \in B_{r_i}$ with $i = 1, \ldots, d$. In other words, we decompose the paths into commonly shared part and unique unshared part, different substructure paths must appear in pair to cancel their unique structures, according to path operations. However, if the unique structures are cancelled, there is no common layers or edges for edge swapping since $E(p_i) \cap E(p_r) = \emptyset$. So no paths in $\cup_{r=1}^{|P_{ind}^S|} B_r$ such that $p - p_{s,1} + p_{r_1,1} - p_{r_1,2} + \cdots p_{r_i,1} - p_{r_i,2} + \cdots + p_{r_d,1} - p_{r_d,2} = 0$ holds. Hence, path set $B = \cup_{r=1}^{|P_{ind}^S|} B_r$ is path independent.

Next we will prove that any path $p \in P$ can be represented by independent path set $B$. For any path $p$ with the structure of $p_r \in P_{ind}^S$ but $p \notin B_r$, it is trivia that path $p$ can be represented by $B_r$ because $B_r$ is the basis path set of $G_r$. If $p$ is out of the structure range of $P_{ind}^S$, the structure of $p$ can be represented by $P_{ind}^S$, because $p$ must belong to one substructure of $P^S$ for **Lemma 3**. Since the constraint requires no commonly shared layers between two independent substructure paths, so there is no other substructure path out of the structure range of $P^S$. This is because we need to do finite times of edge swapping to get a new path but this edge swapping is based on the shared layers according to path operations. Therefore, $B$ is a basis path set and $|B| = \bigcup_{r=1}^{|P_{ind}^S|} |B_r|$. According to **Lemma 2**, $|B_r|$ of fully connected graph $G_r$ without skipping edges is $m_r - H_r$, where $m_r$ is the total number of edges and $H_r$ is the total number of hidden nodes in sub-graph $G_r$ corresponding to each structure path $p_r \in P^S$. So $|B| = \sum_{r=1}^{|P_{ind}^S|}(m_r - H_r)$. ∎

Since every substructure path $p_r \in P^S$ represents one type of substructure information about edge skipping or layer passing in graph $G$, so different types of substructures can be combined together but no shared layers exist between two independent substructures.

## 5 Conclusion

In graph theory, we give the definitions of path operation and basis path in regular fully connected neural network. We investigate attractive properties of basis paths when constructing basis paths in the fully connected neural network without edge skipping. Based on this investigation, we propose hierarchical algorithm HBPS to find the basis path set $B$ in fully connected neural network $G$ with the constraint that no shared layers between two independent substructures. This kind neural network is one special version of fully connected neural network with any edge skipping structure. This algorithm decomposes the fully connected neural network into several independent and parallel substructures. Later algorithm HBPS can be extended to find basis path set for regular fully connected neural network without the constraint about shared layers. The research in paper opens the black box about how basis path set helps to achieve superior performance in the training of ReLU networks, and sheds light on how the basis path set works in regular neural network.

## Appendix A

To make the lemmas, theorems and algorithms clear, the main notations in this paper are summarized in Table 1.

Table 1: Notations

| Notation | Description |
|---|---|
| $O_i^l$ | the $i$-th node in $l$-th layer in the neural network |
| $O^l$ | the $l$-th layer in the neural network |
| $(O_i^l, O_{i'}^{l+j})$ | the directed edge between layer $l$ and layer $l+j$, where $1 \leq j \leq L - l$, $1 \leq i \leq |O^l|$ and $1 \leq i' \leq |O^{l+j}|$. |
| $G = (V, E)$ | the neural network graph $G$ with node set $V = O^0 \cup \ldots \cup O^l \ldots \cup O^L$ and edge set $E = \{(O_i^l, O_{i'}^{l+j}), 0 \leq l \leq L - 1$ and $1 \leq j \leq L - l\}$ |
| $O_{i*}^l$ | some node without specified position in the $l$-th layer |
| $p = (O_{i*}^0, O_{i*}^{1'}, \ldots, O_{i*}^{j'} \ldots, O_{i*}^L)$ | path starting from the input layer $O^0$ to the output $O^L$ passing through from several hidden nodes |
| $P = \{(O_{i*}^0, O_{i*}^{1'}, \ldots, O_{i*}^{j'} \ldots, O_{i*}^L) \mid 0 < 1' < 2' \ldots < j' < L\}$ | set consisting of all paths from the input layer to the output layer in network $G$ |
| $B \subseteq P$ | basis path set |
| $m = |E|$ | the number of edges in graph $G$ |
| $H$ | the number of total hidden nodes in the graph $G$ without skipping edges |
| $\alpha_i$ | sub-structure path vector |
| $G(k) = (O^k \cup O^{k+1}, E^k)$, | sub-graph with $E^k = \{e \in G | e$ leaves from $k$-th layer and enters $k+1$-th layer $\}$ and $0 \leq k \leq L-1$ |
| $P_{dir}^{(k)}$ | direct path set in sub-graph $G(k)$ |
| $P_{cross}^{(k)}$ | cross path set in sub-graph $G(k)$ |
| $P_{dir}(O_i^k)$ | direct path set passing through node $O_i^k$ |
| $P_{cross}(O_i^k)$ | cross path set passing through node $O_i^k$ |
| $P(O_i^k)$ | all paths reaching node $O_i^k$ |
| $G(k)'$ | the sub-graph of $G$ from 0-th layer till $k+1$-th layer |
| $m(k)$ | the number of edges in sub-graph $G(k)'$ |
| $H(k)$ | the number of hidden nodes in sub-graph $G(k)'$ |
| $B(k) = P_{dir}^{(k)}(k) \cup P_{cross}^{(k)}$ | path set in $G(k)'$ |
| $G_r$ | sub-graph of $G$ with the same structure information as $p_r \in P_{ind}^S$ |
| $V^S = \{O_{i*}^0, ..., O_{i*}^l, ..., O_{i*}^L\}$ | randomly selected node sub-set for all layers |
| $E^S = \{(O_{i*}^j, O_{i*}^l) \in E | O_{i*}^j \in V^S, O_{i*}^l \in V^S, 0 \leq j < l \leq L\}$ | edge sub-set based on node sub-set $V^S$ |
| $G^S = (V^S, E^S)$ | sub-graph contains structure information of graph $G$ |
| $P^S$ | sub-structure path set of original graph $G$ |
| $G_r = (V_r, E_r)$. | sub-graph $G_r$ contains the same structure information as path $p_r \in P^S$ |
| $B_r$ | independent path set of $G_r$ |